\documentclass[manuscript]{acmart}

\AtBeginDocument{%
  \providecommand\BibTeX{{%
    \normalfont B\kern-0.5em{\scshape i\kern-0.25em b}\kern-0.8em\TeX}}}

\setcopyright{acmcopyright}
\copyrightyear{2021}
\acmYear{2021}
\acmDOI{nn.nnnn/nnnnnnn.nnnnnnn}

\acmConference[RecSys '21, CARS Workshop]{CARS 2.0: Workshop on Context-Aware Recommender Systems (RecSys '21)}{October 2nd, 2021}{Online, Worldwide}
\acmBooktitle{CARS: Workshop on Context-Aware Recommender Systems (RecSys '21), October 2nd, 2021, Online, Worldwide}
\acmPrice{15.00}
\acmISBN{XXX-X-XXXX-XXXX-X/YY/MM}

\usepackage{multirow}
\usepackage{amsmath}
\begin{document}

\title{Modeling Dynamic Attributes for Next Basket Recommendation}

\author{Yongjun Chen}
 \affiliation{
   \institution{Salesforce}
  \country{USA}
 }

\author{Jia Li}
 \affiliation{
   \institution{Salesforce}
  \country{USA}
 }
 
\author{Chenghao Liu}
 \affiliation{
   \institution{Salesforce}
  \country{Singapore}
 }
 
\author{Chenxi Li}
 \affiliation{
   \institution{Salesforce}
  \country{USA}
 }

\author{Markus Anderle}
 \affiliation{
   \institution{Salesforce}
  \country{USA}
 }
 
\author{Julian McAuley}
 \affiliation{
   \institution{UC San Diego}
  \country{USA}
 }
 
\author{Caiming Xiong}
 \affiliation{
   \institution{Salesforce}
  \country{USA}
 }
 
\renewcommand{\shortauthors}{Yongjun Chen, et al.}

\begin{abstract}
Traditional approaches to next-item
and next basket recommendation
typically extract
users' interests based on their past 
interactions
and associated static contextual information 
(e.g.~a user id or item category).
However,
extracted interests can be inaccurate and 
become
obsolete. 
\emph{Dynamic} attributes, 
such as user income changes, item price changes (etc.),
change
over time. Such dynamics can 
intrinsically reflect the evolution of users' interests.
We argue that modeling such 
dynamic attributes 
can
boost
recommendation performance. 
However, properly integrating 
them into 
user interest models 
is challenging 
since attribute dynamics 
can be diverse 
such as time-interval aware,
periodic patterns (etc.),
and they represent users'
behaviors from different perspectives,
which can happen asynchronously
with interactions.
Besides dynamic attributes,
items in each basket 
contain complex interdependencies 
which might be beneficial 
but nontrivial to 
effectively capture.
To address these challenges, 
we propose a novel \textbf{A}ttentive 
\textbf{n}etwork to model 
\textbf{D}ynamic 
\textbf{a}ttributes 
(named \textbf{AnDa}). 
AnDa separately encodes 
dynamic attributes and basket item sequences.
We design a periodic aware encoder to 
allow the model to capture 
various temporal patterns 
from dynamic attributes. 
To effectively learn useful 
item relationships, intra-basket attention 
module is proposed. 
Experimental results on three 
real-world datasets demonstrate 
that our method 
consistently
outperforms the state-of-the-art.

\end{abstract}





\maketitle

\section{Introduction}

Sequential recommendation systems 
have been successfully 
applied to various applications 
such as product recommendation,
food recommendation, music recommendation, etc. 
There are two lines of 
sequential recommendation tasks 
based on different assumptions 
about users' interaction behavior. 
\emph{Next item} recommendation assumes that 
users interact with items sequentially,
so that recommendations can 
be made by modeling the sequential 
semantics of the interaction history~\cite{kang2018self,li2020time,sun2019bert4rec,liu2021contrastive}, 
possibly including additional
static attributes~\cite{zhang2019feature} 
(product category, brand, etc.).
Next \emph{basket} recommendation~\cite{yu2016dynamic,liu2020basconv} 
assumes that users 
interact with multiple items 
during each round
(i.e.,~a basket). 
The goal is to recommend 
a basket of items that a user 
is likely to interact 
with at the next time step. 
In the next basket recommendation task, 
in addition to
the sequential patterns 
underlying
historical interactions, 
items in
each basket 
, and context of users or items often provides useful information.
.

\begin{figure}[htb]
  \centering
  \includegraphics[width=0.8\linewidth]{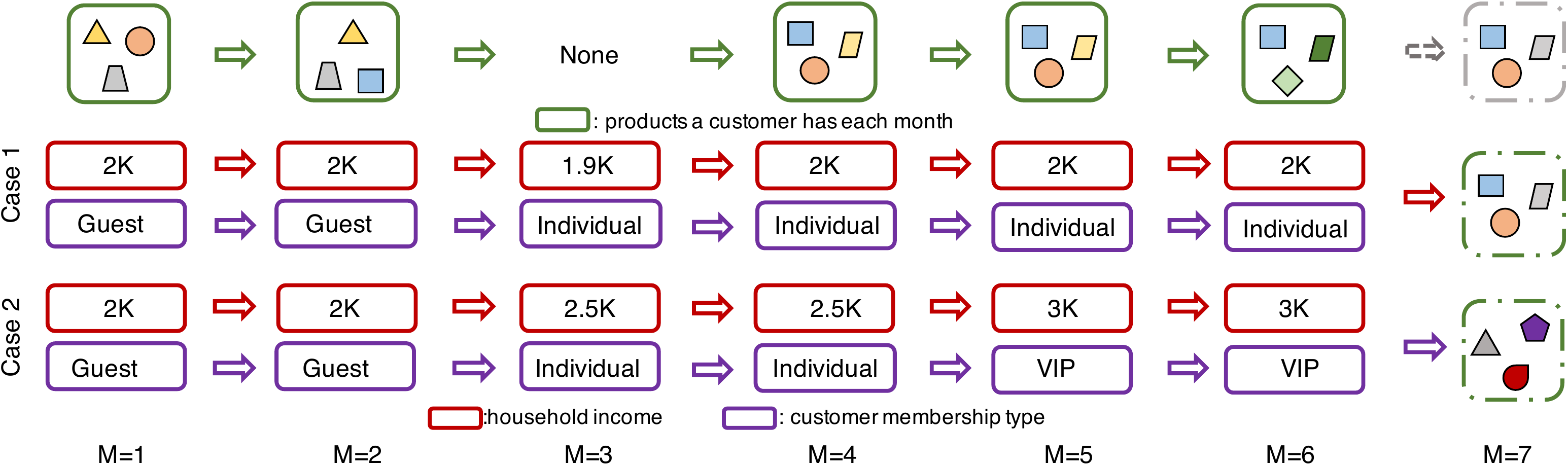}
  \caption{An illustration of a recommender system with or without dynamic attributes.).
  }
  \label{fig:illustration}
\end{figure}

Existing solutions for 
next-basket recommendation tasks train a 
sequential recommender based on the 
users' 
interaction 
history~\cite{rendle2010factorizing,yu2016dynamic,le2019correlation,hu2019sets2sets}, 
or with additional 
static contextual attributes~\cite{bai2018attribute} 
(product category, brand, etc.) to extract 
users' interests. However, 
extracted interests can be inaccurate and obsolete. 
Dynamic attributes, which 
change
over time, 
appear
in many applications, and can provide 
more accurate descriptions of the shift of 
a user's interests or changes in item properties. 
For example, in a bank product recommendation 
scenario (see Figure~\ref{fig:illustration}), 
products from the bank are recommended to customers. 
Given only a sequence of monthly records of 
a customer's
products,
the next basket 
recommended to the user is deterministic.
Instead, dynamic attributes 
such as the household income
and the customer membership type
, which are changing overtime,
%
help a recommender to beter capture a user's changing interests.
Case 1 and 2 in Figure~\ref{fig:illustration} 
illustrate that two customers 
with the same historical purchase behaviors 
but different household income and membership 
type sequences can 
have different interests.

Although it is essential to model dynamic attributes, 
properly integrating dynamic attributes into 
sequential models is challenging.
First, the temporal patterns underlying 
dynamic attributes can be diverse. 
There can be
time-interval (i.e., two items purchased with different time-intervals 
has different impact on users' future purchase behaviors.)
and 
periodic patterns (i.e., seasons, weekday/weekend patterns, etc.).
Second, dynamic attributes represent users' 
behaviors from different perspectives,
and can happen asynchronously with 
interactions.
Directly concatenating
basket items with dynamic attributes at each time
step may not be stable to model 
diverse sequential
patterns.

Besides dynamic attributes, 
in users' historical interactions,
items in each basket contain complex 
interrelationships (item correlations, co-purchases, etc.). 
Existing solutions~\cite{le2019correlation,che2019inter} 
either pre-define static item correlations
based on 
co-occurring items, 
or 
(vanilla) attention to extract
item correlations based on
the whole items the last basket. 
However,
multiple item interrelationships 
can 
exist based on multiple subsets 
of items in a basket and together
influence next basket items. 
For example, in a grocery shopping,
a customer bought apple, banana,  
TV and Speaker. The apple and banana 
are high correlated 
while TV and Speaker are 
also high correlated. The user may 
purchase some fruits again with 
TV accessories.
Using multi-head attention
allows the model to capture
different item relationships
under
different subset of items in a historical 
basket.

To address the above challenges, 
we propose a novel \textbf{A}ttentive \textbf{n}etwork 
to model the \textbf{D}ynamic \textbf{a}ttributes 
as well as users' historical 
interacted items (\textbf{AnDa} for short). 
AnDa separately encodes and learns 
representations from dynamic attributes and
interactions with basket items.
To allow the model to capture time-interval aware 
and periodic patterns, 
we propose an input encoder containing 
a time-aware padding and periodic index embedding 
to encode dynamic attributes. 
To capture complex 
item relationships in each basket,
an intra-basket attentive module is introduced. 
It is applied to each basket item to 
extract useful item relationships. 
We conduct experiments on three real-world 
datasets for 
next basket recommendation. 
Our
experimental results demonstrate that our proposed method
significantly outperforms baseline approaches.

\section{Related Work}
\label{sec:related}

\subsection{Next Basket Recommendation}

To capture sequential patterns at a basket level for 
next basket recommendation, 
DREAM~\cite{yu2016dynamic} encodes items in each basket 
using 
max and average pooling and learns 
the sequence representation through 
an RNN-based network. 
ANAM \cite{bai2018attribute} 
improves upon DREAM by considering 
static item attributes 
using vanilla attention. 
Sets2Sets \cite{hu2019sets2sets} 
views the task as a multiple baskets prediction 
and proposes a RNN based encoder-decoder method 
to improve the performance. 
To capturing item relationships 
at each basket,
Beacon~\cite{le2019correlation} 
defines 
pre-computed static item correlation 
information based on the 
co-occurring items 
in the observed training baskets 
and then incorporates it into 
an RNN-based model. 
IIANN~\cite{che2019inter} learns the 
correlation between 
the most recent basket items 
and the target item 
to summarize 
users' short-term interests.
In this work, we
use multi-head self-attention
within each basket so that
complex item interrelationships
(e.g., co-occurrences) can 
be captured. MITGNN~\cite{liu2020basket}
focus on capturing users' intention information
in each basket and across different users'.
In comparison, our work focus on 
leverage dynamic attributes for providing 
more accurate user interests.

\vspace{-0.2cm}
\subsection{Feature Interaction Learning}
Factorization Machines (FMs)~\cite{rendle2010factorization,cheng2014gradient,juan2016field,xiao2017attentional}
capture
second-order feature interactions and 
have proven to be effective for 
recommendation~\cite{rendle2011fast}. 
With the success of deep neural networks (DNN),
many works start to explore high-order
feature interactions using DNN.
NFM~\cite{he2017neural} combines 
FM with a DNN to model 
high-order feature interactions. 
The Wide\&Deep~\cite{cheng2016wide} 
model uses a wide part to model 
second-order interactions 
and a deep part to 
model the higher-order interactions. 
Different from Wide\&Deep, 
Deep\&Cross~\cite{wang2017deep} 
uses a cross-product transformation 
to integrate features, and 
xDeepFM~\cite{lian2018xdeepfm} 
proposes a CIN module to
take the outer product 
at a vector-wise level. 
These works, which use DNN to 
capture high-order feature interactions
implicitly,
lack good explanation ability in general.
To this end, AutoInt~\cite{song2019autoint} 
uses a self-attention mechanism 
to model high-order interactions 
with a more precise explanation 
of the interacted features. 
Inspired by this, we also apply 
multi-head self-attention to learn 
higher-order feature interactions 
and item interrelationships in each basket.

\subsection{Attention Mechanisms}


With the success of 
Transformer networks 
in machine translation 
tasks~\cite{vaswani2017attention}, 
purely attention-based models 
SASRec~\cite{kang2018self} is 
the first work that 
uses a pure self-attention mechanism 
to model sequential recommendation 
and demonstrates better performance 
than RNN-based methods. 
TiSASRec~\cite{li2020time} 
extends SASRec using self-attention 
to model the time interval 
between two adjacent interactions. 
BERT4Rec~\cite{sun2019bert4rec} uses 
bidirectional self-attention 
with the Cloze objective. 
FDSA~\cite{zhang2019feature} 
further improves 
sequential recommendation 
by incorporating the usage of 
static item attributes 
and using vanilla attention 
to capture users' interests. 
However, it does not 
incorporate dynamic user attributes 
and only models the sequential patterns 
at the item level 
instead of the basket level. 
Although this approach 
could be extended for 
modeling dynamic features, 
it uses vanilla attention 
to average out different attributes 
at each time step. 
As a result, 
it is not able to 
learn high-order feature interactions. 
Also, it loses the temporal aspects of 
individual features. 
For example, 
if the trend of 
a usage metric is going up or down 
in the past three months, 
FDSA will not capture 
such a trend 
due to its averaging operation. 
Instead, our time-level attention module 
can capture these temporal patterns.

\section{Proposed Approach}
\label{sec:model}

\setlength{\tabcolsep}{0.9mm}{
\begin{table}[tb]
  \caption{Summary of Main Notation.}
  \label{tab:notations}
  \begin{tabular}{cc||cc}
    \toprule
    Notation & Description & Notation & Description\\
    \midrule
    $U$ & set of users with total number $|U|$ & $A_{t, \mathit{num}}^{u}$ & numerical attributes values of $u$ at time $t$\\
    $V$ & set of items with total number $|V|$ & $e^{V}_{t}$, $e^{F_{\mathit{cat}}}_{t}$ & multi-hot vector of $B_{t}^{u}$ and $A_{t, \mathit{cat}}^{u}$\\
    $F$ & set of attributes with total number $|F|$ & $p_{t}$ & positional embedding vector at time $t$\\
    $B_{t}^{u}$ & a basket of items that $u$ interacted with at time $t$ & $m_{t}$ & periodic index embedding vector at time  $t$\\
    $A_{t, \mathit{cat}}^{u}$ & categorical attribute values of $u$ at time $t$ & $r_{i,t}$ & predicted relevance score for item $i$ at time $t$\\
  \bottomrule
\end{tabular}
\end{table}}

\begin{figure*}[htb]
\setlength{\abovecaptionskip}{0.05cm} 
  \centering
  \includegraphics[width=0.9\linewidth]{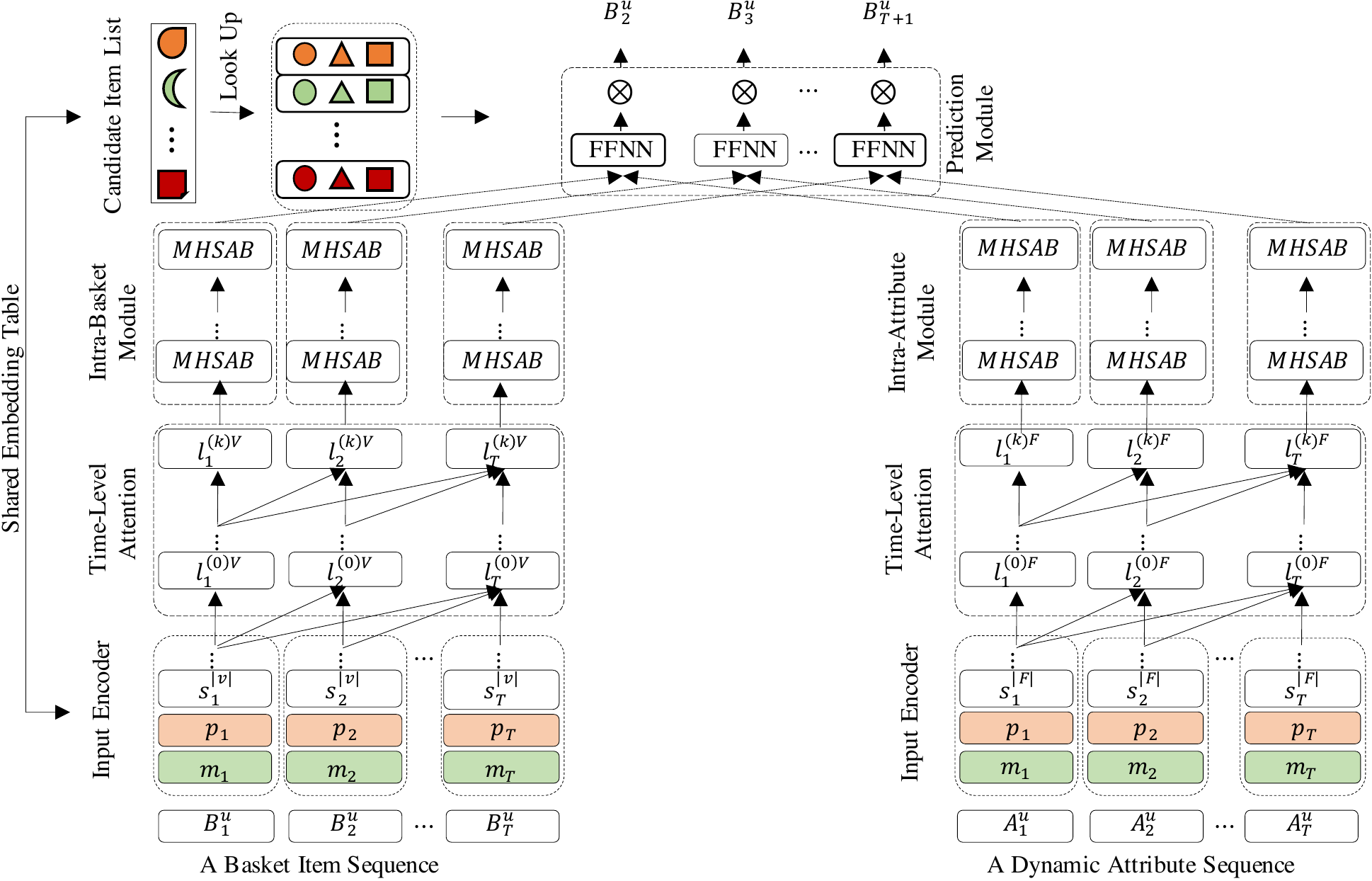}
  \caption{The Model Architecture of AnDa.}
  \label{fig:model-architecture}
\end{figure*}

In this section, 
we first define the notation 
and formalize 
the next basket recommendation task 
with dynamic user attribute information. 
And then we present the proposed framework \textbf{AnDa} in detail. 
The framework is illustrated in Figure~\ref{fig:model-architecture}
and the notation is summarized 
in Table~\ref{tab:notations}.

\vspace{-0.2cm}
\subsection{Problem Statement}
In next basket recommender systems with dynamic attributes, 
historical 
basket
interactions
and dynamic attribute sequences are given, 
and the goal is 
to recommend 
the next basket's items.
Formally, 
we denotes $U$, $V$ 
and $F$ a sets of users, 
items and user attributes respectively.
For a user $u \in U$, a sequence of baskets 
$B^{u}=\left \{ B_{1}^{u},B_{2}^{u},\cdots ,B_{T_{u}}^{u} \right \}$ represents his or her item interactions 
sorted by time. 
$T$ is the maximum time steps 
of each sequence, and
$B_{t}^{u} \subseteq V$ is a set of items 
that user $u$ interacted with at time step $t$. 
A sequence 
$A^{u}=\left \{ A_{1}^{u},A_{2}^{u},\cdots ,A_{T_{u}}^{u} \right \}$ represents the value of 
dynamic user attributes 
of user $u$ ordered by time. 
Specifically, $A_{t}^{u} \in F$ 
are all the attribute values 
of $u$ at time step $t$. 
The goal is to predict basket items 
that user $u$ will interact with 
at time step $T+1$ given 
$T$ 
historical 
baskets $B^{u}$ and attributes
$A^{u}$.


\subsection{Time-Interval and Periodic Aware Input Encoder}

\textbf{Embedding Lookup:} For each basket 
of dynamic attributes $A_{t}^{u}$,
we model categorical 
and numerical features differently. 
Categorical attributes 
$A_{t, \mathit{cat}}^{u} \subseteq A_{t}^{u}$ 
are represented 
by an $|F_{\mathit{cat}}|$-dimensional 
multi-hot vector 
denoted by  
$e_{t}^{F_{\mathit{cat}}}\in \mathbb{R}^{|F_{\mathit{cat}}|\times 1}$. 
Numerical attributes 
are normalized 
into the range $\lbrack -1, 1 \rbrack$ 
using 
min-max normalization, 
denoted as 
$s_{t}^{F_{\mathit{num}}}\in \mathbb{R}^{|F_{\mathit{num}}| \times 1}$.
Each basket of items 
$B_{t}^{u}$ is represented by 
a $|V|$-dimensional 
multi-hot representation, 
denoted by $e_{t}^{V}\in \mathbb{R}^{|V|\times 1}$. 
After that, we apply
a
concatenation-based 
lookup function~\cite{bai2018attribute} to encode 
$e_{t}^{F_{\mathit{cat}}}$ and $e_{t}^{V}$:
\begin{equation}
  s_{t}^{F_{\mathit{cat}}} = \mathit{CONCAT-LOOKUP}(R, e_{t}^{F_{\mathit{cat}}}), s_{t}^{V} = \mathit{CONCAT-LOOKUP}(Q, e_{t}^{V})
\end{equation}





where 
$R \in \mathbb{R}^{|F_{\mathit{cat}}| \times D}$ 
and $Q \in \mathbb{R}^{|V| \times D}$ 
are learnable embedding matrices 
for categorical attributes and items. 
The ``CONCAT-LOOKUP'' function 
retrieves the corresponding 
embedding vectors 
and then concatenates them together 
to form matrices $s_{t}^{F_{\mathit{cat}}} \in \mathbb{R}^{|F_{\mathit{cat}}| \times D}$, 
and $s_{t}^{V} \in \mathbb{R}^{|V_{t}| \times D}$. $D$ 
is the embedding dimension of 
each item and categorical attribute. 
$|V_{t}|$ is the total number of items 
in $B_{t}^{u}$. 
Since the number of items 
in each basket varies, 
we set the maximum number of items 
in the basket as the largest basket size 
in the training set $|V_{\mathit{max}}|$, 
and add padding vectors for baskets
smaller
than $|V_{\mathit{max}}|$.

\textbf{Time-aware Padding Operation:} 
We set the maximum sequence length 
as $T$ to get up to 
the latest $T$ position steps' information. 
If the sequence length 
is shorter than $T$, 
a zero-pad operation 
will be applied to 
empty positions. 
Otherwise, 
we truncate to the 
last
$T$ positional steps.
Unlike previous 
works~\cite{kang2018self,zhang2019feature,sun2019bert4rec} 
that pad zeros 
to the left 
until the sequence length is $T$, 
we pad zeros 
to the missed time steps 
to keep the time interval information. 
We denote 
$S^{V}$, 
$S^{F_{\mathit{cat}}}$, and 
$S^{F_{\mathit{num}}}$ 
as the padded basket item, 
categorical attribute, 
and numerical attribute sequences 
respectively.


\textbf{Periodic Index Embedding:} 
We introduce 
a periodic index embedding 
$M\in \mathbb{R}^{T'\times D}$ 
for attention modules 
to discover periodic patterns. 
The index repeats over 
every $T'$ time steps of a sequence. 
For example, $T'=12$ can be used for 
capturing seasonal patterns 
when the time interval 
between each two baskets is one month. 
Positional embeddings~\cite{shaw2018self}  that commonly used
to identify item positions is also used 
in this paper.
Formally, 
we concatenate the periodic 
and positional index embedding 
with $s_{t}^{V}$ as 
$l_{t}^{V} = [s_{t}^{V}, p_{t}, m_{t}]$,
where $t \in \{1,\cdots, T\}$, 
$m_{t} \in M$ and $p_{t}\in P$ are 
$D$ dimensional periodic 
and positional embedding vectors. 
Then a basket sequence is represented as $L^{V}=\left \{ l_{1}^{V},l_{2}^{V},\cdots ,l_{T}^{V} \right \}$. We also add positional and periodic index embeddings to $S^{F_{\mathit{cat}}}$ and $S^{F_{\mathit{num}}}$ to form $L^{F_{\mathit{cat}}}$ and $L^{F_{\mathit{num}}}$ respectively.

\vspace{-0.2cm}
\subsection{Time Level Attention Module}

To capture temporal patterns 
from $L^{V}$, $L^{F_{\mathit{cat}}}$ 
and $L^{F_{\mathit{num}}}$,  
we separately encoder them via
multi-head self-attention (MHSA)~\cite{vaswani2017attention}. 
Formally, let $L^{0}=L=L^{F_{\mathit{cat}}}$, and then fed
into a MHSA block as shown below:
Eq.~\ref{eq:mult-head-self-attention}
\begin{equation}\label{eq:mult-head-self-attention}
  \begin{aligned}
    M^{(1)} = \mathit{MHSA}(L^{(0)}, h) = [\mathit{head_{1}}, \mathit{head_{2}}, \ldots, \mathit{head_{h}]W_{\mathit{Concate}}},\\
    \mathit{head_{i}} = SA_{i}(L^{(0)}W_{\mathit{Query}}^{{i}}, L^{(0)}W_{\mathit{Key}}^{{i}}, L^{(0)}W_{\mathit{Value}}^{{i}}),\\
  \end{aligned}
\end{equation}

where $h$ is 
the number of sub-spaces, 
$W_{\mathit{Query}}^{{i}}\in \mathbb{R}^{C\times C}$, $W_{\mathit{Key}}^{{i}}\in \mathbb{R}^{C\times C}$, 
$W_{\mathit{Value}}^{{i}}\in \mathbb{R}^{C\times C}$ and $W_{\mathit{Concate}}\in \mathbb{R}^{C'\times C}$ 
are learned parameters 
($C=(|F_{\mathit{cat}}+2|\cdot D)$, and $C'=hC$). Similar, $L^{V}$ and $L^{F_{\mathit{num}}}$ 
are also encoded via Eq.~\ref{eq:mult-head-self-attention}.

Following~\cite{kang2018self}, 
we add causality mask to avoid future information leek.
To enhance the representation learning 
of the self-attention block, 
residual connections~\cite{he2016deep}, dropout~\cite{srivastava2014dropout}, 
layer normalization~\cite{ba2016layer}, and 
two
fully connected layers with 
ReLU activation functions 
are added to 
form the entire 
multi-head self-attention block (MHSAB) 
as follows:

\begin{equation}
  L^{(1)} = \mathit{MHSAB}(L^{(0)}, h)
\end{equation}
We stack multiple attention blocks 
to capture more complex feature interactions:
\begin{equation}\label{eq:mhsab}
  L^{(k)} = \mathit{MHSAB}(L^{(k-1)}, h^{k}), k > 0
\end{equation}
where $L^{(0)}=L$ and $h^{k}$ 
is the number of heads 
at the $k^{\mathit{th}}$ attention block. 
$L^{(k)}$ 
is the output 
after stacking multiple time-level attention
layers.
The extracted representation vector 
at time step $t$ can be denoted 
$L_{t}^{(k)}$ and 
contains 
information extracted from time 
$1$ to time $t$ of the input sequence. 

\subsection{Intra-Basket and Intra-Attribute Self-Attention Modules}

Item correlations in each basket can 
reveal
some useful information such as 
co-purchase relationships. 
The key problem is 
how to determine which items 
should be combined or are correlated.
In this paper, 
we use multi-head self-attention 
to learn information 
such as items' correlation relationships.
Specially, 
given representations of all items 
in a basket $L_{t}^{k}$, 
a single-head self-attention module 
will first compute the similarity matrix
which is seen as item correlation scores, 
and then it updates the item representation $l_{t}^{k}$ 
by combining 
all relevant items using the similarity coefficients 
(generated based on the similarity matrix). 
We use MHSAB 
to enhance the model's 
capability of capturing 
complex item correlations, 
which is formed as follows:

\begin{equation}\label{eq:intra-attention}
  L_{t}^{(k+1)} = \mathit{MHSAB}(L_{t}^{(k)}, h^{k+1}), k >0, t \in \{1,2,\cdots,T\}
\end{equation}
where  $L_{t}^{(k)}$ is the output of the time level attention module at time $t$, and $h^{k+1}$ is the $(k+1)^{\mathit{th}}$ attention block. 

We stack Eq.~\ref{eq:intra-attention} $m$ 
times to 
capture
more complex item relationships $L_{t}^{(k+m)V}$.
Similarly, we can stack Eq.~\ref{eq:intra-attention} on dynamic attribute (named Intra-Attribute Attention)
to get higher level categorical attribute 
interactions $L_{t}^{(k+m)F_{\mathit{cat}}}$.

\subsection{Model Training}
The encoded user representations are projected as:
$L_{t}^{\mathit{all}} = \mathit{FFNN}([L_{t}^{(k+m)F_{V}}, L_{t}^{(k+m)F_{\mathit{cat}}}, L_{t}^{(k+m)F_{\mathit{num}}}])$, where $\mathit{FFNN}$ is a feed forward network, $L_{t}^{\mathit{all}} \in \mathbb{R}^{1\times D}$ 
is the final representation 
given dynamic attributes 
and basket items from time step $1$ to $t$. 
We then adopt the binary cross-entropy loss as the objective 
for training the model defined as:
\begin{equation}
  -\sum_{B^{u}\in B}\sum_{t\in \{1,\cdots,T\}]}[\sum_{i\in B^{u}_{t+1}}log(\sigma (L^{\mathit{all}}_{t}\cdot Q_{i})))+\sum_{j\notin B^{u}_{t+1} }log(1-\sigma (L^{\mathit{all}}_{t}\cdot Q_{j}))]
\end{equation}
where $\sigma$ is the 
sigmoid function $\sigma (x) = 1 /(1+e^{-x})$.
$Q$ is the item embedding matrix 
which is shared for 
encoding basket items in input encoders.
The target basket items for user 
$u$ are a shifted version of $B^{u}$, 
denoted by 
$\left \{ B_{2}^{u},B_{3}^{u},\cdots ,B_{T+1}^{u} \right \}$.

\section{Experiments}
\label{sec:experiments}


\subsection{Experimental Setting}
\subsubsection{Datasets}
Table \ref{tab:dataset-information} 
summarizes the statistics of the datasets. 
EPR is a private dataset 
sampled from a leading enterprise cloud platform. 
The task is to recommend products to businesses. 
Examples of the dynamic attributes 
are 
behavior metrics on the website, 
sales information, 
and marketing activities of the business.
SPR\footnote{https://www.kaggle.com/c/santander-product-recommendation} is a public dataset 
on product recommendations 
for the Santander bank. 
Ta-Feng\footnote{https://www.kaggle.com/chiranjivdas09/ta-feng-grocery-dataset} is a grocery shopping dataset. 
We followed Beacon~\cite{le2019correlation} 
to create train, validation, and test sets 
by chronological order. 
The <train, validation, test> sets 
for EPR, SPR, and Ta-Feng datasets are
<$1^{\mathit{st}}$-$20^{\mathit{th}}$, $21^{\mathit{st}}$, $22^{\mathit{nd}}$-$24^{\mathit{th}}$>, 
<$1^{\mathit{st}}$-$16^{\mathit{th}}$, $17^{th}$, $18^{\mathit{th}}$> , 
and <$1^{\mathit{st}}$-$3^{\mathit{rd}}$, 
successive 0.5, last 0.5> 
month(s) respectively. The interval between each two time steps is 1 month for EPR and SPR datasets, and 1 day for Ta-Feng dataset.

\begin{table}[htb]
  \caption{Dataset information.}
  \label{tab:dataset-information}
  \begin{tabular}{ccccccc}
    \toprule 
    Datasets/Information      & \# Users & \# Items & \# Attributes & Sparsity & Avg. Baskets & Avg. Basket Size\\
    \midrule
    EPR         & 229314 & 23 & 169 & 7.92\% & 17.04 & 1.34 \\
    SPR         & 956645     & 24     & 24 & 4.92\% & 14.51 & 1.47 \\
    Ta-Feng    & 13541    & 7691     & 5 & 3.58\% & 7.12 & 5.8\\
    
  \bottomrule
\end{tabular}
\end{table}
\subsubsection{Baselines \& Evaluation Metrics}
We include three groups of baseline methods: PopRec 
considers no
sequential patterns; FMC~\cite{rendle2010factorizing} 
and FPMC~\cite{rendle2010factorizing} are Markov Chain-based 
sequential methods; and Neural Network based methods 
with (FDSA+) or without (DREAM~\cite{yu2016dynamic}, 
Beacon~\cite{le2019correlation}, 
Sets2sets~\cite{hu2019sets2sets}, 
and CTA~\cite{wu2020deja}) 
dynamic attributes.
We evaluate all methods on the whole item set 
without sampling. 
All the items are first ranked and then evaluated by
Hit Rate ($\mathit{HR}@K$), Normalized Discounted 
Cumulative Gain ($\mathit{NDCG}@K$), and Mean Average 
Precision ($\mathit{MAP}$). In this work, we report HR 
and NDCG with K=5.


\subsubsection{Parameter Settings}
We tune the embedding dimension $C$ from \{10, 15, 30, 50\}, 
learning rate from \{0.0001, 0.001, 0.01\}, 
and dropout from \{0.0, 0.1, 0.2, 0.5\}. 
For DREAM, we tune with RNN, GRU, and LSTM modules. 
AdamOptimizer is used to 
update the network with moment estimates 
$\beta _{1}=0.9$ and $\beta _{2}=0.999$. 
For AnDa, we tune the self-attention layers from $\{1,2,4\}$ 
and head number on each attention block 
from $\{1,2,4,6\}$. Maximum sequence lengths are 
set as 12, 16, and 30 in EPR, SPR, and Ta-Feng
respectively.
We report the hyper-parameter sensitivity study results in Figure~\ref{fig:ablation-study-and-hpo}.

\subsection{Overall Performance Comparison}
\setlength{\tabcolsep}{0.9mm}{
\begin{table*}
  \caption{Performance Comparison of different 
  methods on next basket recommendation. 
  Bold/underlined scores are the best/second best in each row.
  The last column 
  shows
  AnDa's relative 
  improvement over the best baseline.
  }
  \label{tab:ptb-spr-ranking-result}
  \begin{tabular}{cccccccccccccc}
    \toprule
    Dataset & Metric & FMC & FPMC & PopRec  & CTA & Sets2Sets & DREAM  & Beacon & FDSA+ & AnDa &  Improv.\\
    \midrule
    \multirow{4}{*}{EPR} & HR@5  & 0.1024 & 0.1297 & 0.4099 & 0.3108 & 0.4155 & 0.2285 & 0.3211 & \underline{0.4613} & \textbf{0.5622} & 
                         21.87\%\\
                         & NDCG@5 & 0.0691 & 0.0889 & 0.2189 & 0.2065 & 0.2207 & 0.1082 & 0.1453 & \underline{0.3211} & \textbf{0.4616} &
                         43.76\%\\
                         & MAP    & 0.0938 & 0.1103 & 0.1805 & 0.1789 & 0.1912 & 0.1324 & 0.1385 & \underline{0.2556} & \textbf{0.3183} &
                         24.53\%\\
    \midrule
    \multirow{4}{*}{SPR}     & HR@5    & 0.2197 & 0.3246 & 0.4527 & 0.4783 & 0.5632 & 0.1441 & 0.5027 & \underline{0.6834} & \textbf{0.7481} & 9.47\%\\
                             & NDCG@5  & 0.1026 & 0.1358 & 0.1192 & 0.2137 & 0.3021 & 0.0694 & 0.2259 & \underline{0.3123} & \textbf{0.4049} & 29.65\%\\
                             & MAP     & 0.1196 & 0.1271 & 0.1465 & 0.1842 & 0.2365 & 0.0991 & 0.1699 & \underline{0.2476} & \textbf{0.2638} & 6.54\%\\

    \midrule
    \multirow{4}{*}{Ta-Feng} & HR@5    & 0.0064 & 0.0089 & 0.0414 & 0.0379 & \underline{0.0498} & 0.0401 & 0.0442 & 0.0301 & \textbf{0.0573} & 15.06\%\\
                             & NDCG@5  & 0.0035 & 0.0044 & 0.0155 & 0.0214 & \underline{0.0271} & 0.0226 & 0.0256 & 0.0141 & \textbf{0.0308} & 13.65\%\\
                             & MAP     & 0.0027 & 0.0039 & 0.0229 & 0.0202 & \underline{0.0259} & 0.0200 & 0.0255 & 0.0188 & \textbf{0.0265}  & 2.32\%\\
    \midrule
    \midrule
    \multirow{1}{*}{Inference Time} & msec./seq. & \textbf{0.1697} & \underline{0.1929} & 0.8428 & 1.8341 & 2.3212 & 2.2542 & 0.6911 & 1.9521 & 2.5172 & -\\

  \bottomrule
\end{tabular}
\end{table*}}

Table~\ref{tab:ptb-spr-ranking-result} 
shows overall results compared with 
baseline approaches. We observe that, first, 
our proposed approach consistently outperforms all 
baselines significantly in terms of Hit Rate, 
NDCG, and MAP by 3.65\% - 21.87\%, 
9.09\% - 43.76\%, and 2.32\% - 24.53\%, 
which demonstrate the effectiveness of our proposed method.
Beside,
The next basket recommenders (DREAM, Beacon, CTA, 
and Sets2Sets) outperform 
those
for 
next item recommendation (FMC, FPMC). 
This indicates that learning the 
sequential patterns with the encoding of 
the intra-baseket information can better 
capture users' dynamic interests. 
The FDSA+ method performs the best among 
baselines in the EPR and SPR datasets, 
while Sets2Sets performs the best in the 
Ta-Feng dataset. The main reason is that 
FDSA+ leverages 
attribute information where EPR and SPR have 
more 
attributes. 
We also report the models’ average inference time 
(milliseconds per sequence) on 400 sequence inputs in 
Table~\ref{tab:ptb-spr-ranking-result} 
(last row). The proposed method takes more 
time to generate recommendation lists 
than baseline methods, 
though is
comparable 
with Set2Sets, DREAM, CTA, and FDSA+.

\subsection{Ablation Study}

To understand the impact of different components in AnDa, 
we conduct a detailed ablation study 
using the SPR dataset in Table \ref{tab:ptb-ranking-result-abl-study}. 
\textbf{AnDa(P)} is AnDa without periodic 
index embedding. The results show that 
the periodic index can help capture 
users' seasonal purchase patterns, 
and thus helps to improve 
performance. \textbf{AnDa(B)} is 
AnDa with basket information only. Without 
dynamic attributes, \textbf{AnDa(B-)} removes the intra-basket module 
from AnDa(B),
\textbf{AnDa(T)} is AnDa 
without using intra-basket and 
intra-attribute modules on both items 
and attributes, and \textbf{AnDa(I)} 
is AnDa without applying the time level 
attention module.
The performance 
degradation on the sub-models shows 
the benefits of each component. 

\begin{table*}[htb]
\caption{Ablation Study on the SPR Dataset.}
\label{tab:ptb-ranking-result-abl-study}
\begin{tabular}{c|cccc|c|cccccc}
\toprule
Models & HR@5 & NDCG@5 & MAP & Average & Models & HR@5 & NDCG@5 & MAP & Average \\ 
\hline
FDSA(+)  & 0.6834           & 0.3123         & 0.2476 & -13.91\% & AnDa(P)  & 0.7340      & 0.3961        & 0.4749   & -1.42\% \\
AnDa(B)  & 0.7408             & 0.4037       & 0.2605 & -0.90\% & AnDa(B-) & 0.7280       & 0.3933        & 0.4753   & -1.89\%\\ 
AnDa(T)  & 0.7401         & 0.3977       & 0.2610 & -1.06\% & AnDa(I)  & 0.7182   & 0.3813        & 0.4666     & -3.61\%\\ 
\hline
AnDa     & \textbf{0.7481}         & \textbf{0.4049}    & \textbf{0.2638} & - & AnDa     & \textbf{0.7481}         & \textbf{0.4049}    & \textbf{0.2638} & - \\ 
\bottomrule
\end{tabular}
\end{table*}


\vspace{-0.4cm}
\subsection{Attention Visualization}
\begin{figure}[htb]
  \centering
  \includegraphics[width=0.8\linewidth]{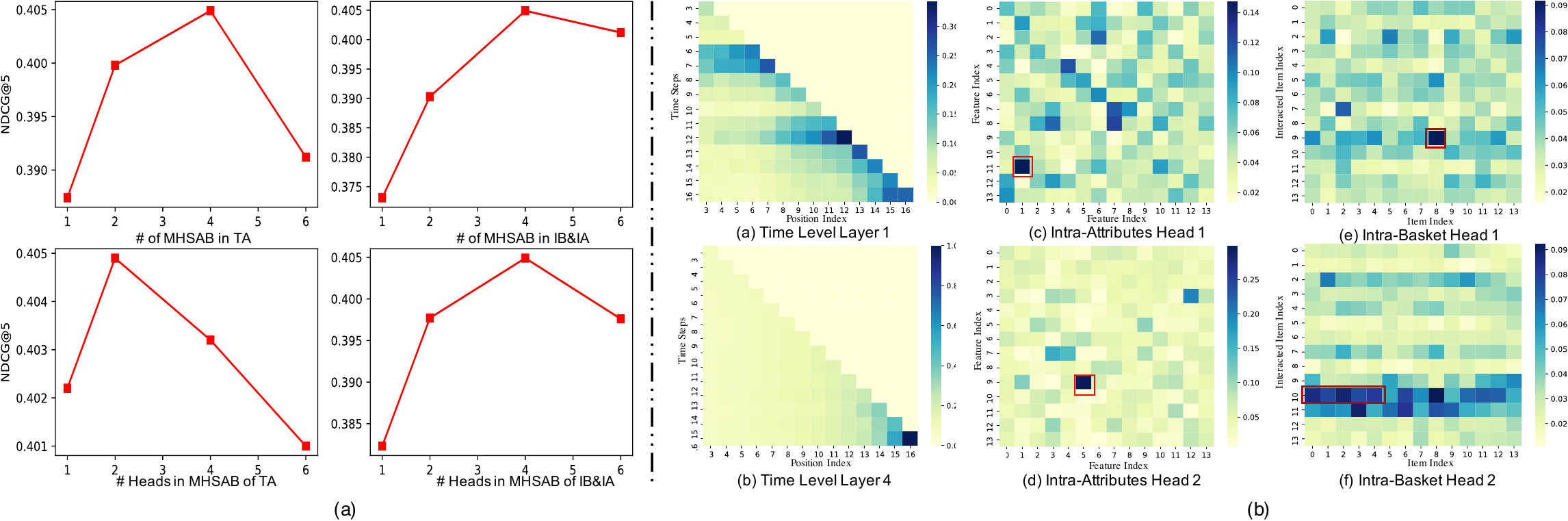}
  \caption{(A): hyperparamter sensitivity study results of AnDa. 
  (B): 
  visualization of attention weights on different MHSA modules.}
  \label{fig:ablation-study-and-hpo}
\end{figure}

We visualize the attention weights of time-level,
intra-attribute, 
and intra-basket attentions 
on sampled sequences from 
the SPR dataset in Figure \ref{fig:ablation-study-and-hpo} (B) 
to gain more insights. 
(a) and (b) are attention weights 
from two different layers (layer 1 and 4) 
of time level basket attention, 
(c) and (d) are from two different heads of 
the first intra-attribute layer,
and (e) and (f) are from two different head of
the first intra-basket layer. 
From (a) and (b), 
we can see the attention 
varies over different layers. 
While the weights in layer 4 
focus more on recent items, 
the weights in layer 1 attend more 
evenly to all previous histories. 
From (c) and (d), 
we observe that the attention weights 
vary over different heads, 
and the module captures meaningful 
feature interactions. 
For example, in (c), 
the position $(11, 1)$ 
(marked by a red square) 
corresponding to interacted feature value 
<\emph{"Foreigner index": NO}, 
\emph{"Customer's Country residence": ES}> 
(the bank is based in Spain, 
so a customer who lives in Spain 
is not a foreigner). 
We can also observe
that intra-basket attention
can capture different item relationships 
under different heads comparing with (e) and (f).

\vspace{-0.3cm}
\section{Conclusion}
\label{sec:conclusion}

In this paper, 
we propose a novel attentive network AnDa, 
which models dynamic attributes 
to better capture
users' dynamically 
changing interests and intentions. 
AnDa separately extracts 
temporal patterns 
from dynamic attributes 
and user historical 
interactions with a novel input encoder.
AnDa
also generates feature interactions and uncovers
item interrelationships 
in each basket with proposed
intra-attribute and 
intra-basket modules respectively.
We evaluate AnDa 
on three real-world datasets and demonstrate 
the usefulness of modeling dynamic attributes 
for next basket recommendation.

\bibliographystyle{ACM-Reference-Format}
\bibliography{anda}
\appendix

\end{document}